\pdfoutput=1

\documentclass[11pt]{article}

\usepackage[final]{acl}

\usepackage{times}
\usepackage{latexsym}

\usepackage[T1]{fontenc}

\usepackage[utf8]{inputenc}

\usepackage{microtype}

\usepackage{inconsolata}

\usepackage{graphicx}

\usepackage{bm} 
\usepackage[usenames,dvipsnames, table]{xcolor}
\usepackage{hyperref} 

\usepackage{enumitem}
\setlist{leftmargin=5.5mm}
\usepackage{linguex}
\usepackage{amsmath}
\usepackage{amssymb}
\usepackage{pifont}
\usepackage{ntheorem}
\usepackage{wrapfig}
\usepackage{cleveref}
\usepackage{todonotes}
\usepackage{booktabs}
\usepackage{multirow}
\usepackage{subcaption}
\usepackage{soul}

\usepackage{CJKutf8}

\newcommand{\mathcolour}{PineGreen}

\newtheorem{pred}{Prediction}
\newtheorem{subresearchq}{Sub Research Question}
\theoremseparator{:}

\def\defn#1{\textbf{#1}}

\title{Modeling Bottom-up Information Quality during Language Processing}

\author{Cui Ding \\
    University of Zürich \\
    \href{mailto:cui.ding@uzh.ch}{\texttt{cui.ding@uzh.ch}} \\\And
    Yanning Yin \\
    University of Basel\\
    \href{mailto:yanning.yin@unibas.ch}{\texttt{yanning.yin@unibas.ch}}\\\AND
    Lena A. Jäger \\
    University of Zürich \\
    \href{mailto:lenaann.jaeger@uzh.ch}{\texttt{lenaann.jaeger@uzh.ch}} \\\And
    Ethan Gotlieb Wilcox \\
    Georgetown University \\
    \href{mailto:ethan.wilcox@georgetown.edu}{\texttt{ethan.wilcox@georgetown.edu}}
    }

\begin{document}
\maketitle
\begin{abstract}

Contemporary theories model language processing as integrating both top-down expectations and bottom-up inputs.
One major prediction of such models is that the quality of the bottom-up inputs modulates ease of processing---noisy inputs should lead to difficult and effortful comprehension.
We test this prediction in the domain of reading.
First, we propose an information-theoretic operationalization for the ``quality'' of bottom-up information as the mutual information (MI) between visual information and word identity.
We formalize this prediction in a mathematical model of reading as a Bayesian update.
Second, we test our operationalization by comparing participants' reading times in conditions where words' information quality has been reduced, either by occluding their top or bottom half, with full words.
We collect data in English and Chinese.
We then use multimodal language models to estimate the mutual information between visual inputs and words.
We use these data to estimate the specific effect of reduced information quality on reading times.
Finally, we compare how information is distributed across visual forms.
In English and Chinese, the upper half contains more information about word identity than the lower half.
However, the asymmetry is more pronounced in English, a pattern which is reflected in the reading times.

\vspace{.2em}
\hspace{1.25em}\includegraphics[width=1.25em,height=1.25em]{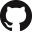}{\hspace{.75em}\parbox{\dimexpr\linewidth-2\fboxsep-2\fboxrule}{\url{https://github.com/DiLi-Lab/Bottom-Up-Information.git}}}

\end{abstract}

\section{Introduction}

During reading, individuals actively expend cognitive effort to extract information. 
Many contemporary theories of language comprehension in general, and reading in particular, model this process as a rational integration of bottom-up and top-down information \citep{legge1997mr, norris2006bayesian, bicknell2010rational, gibson2013channel, gauthier-levy-2023-neural}.
Bottom-up information refers to the perceptual input (e.g., visual forms of words), while top-down information includes the prior beliefs and expectations about what messages or word-forms are likely to be encountered, and is guided by the reader's linguistic and contextual knowledge.
A central prediction of such models is that the ease of reading should be influenced by the quality of the bottom-up information.
In the modality of visual reading, visual signals that effectively convey information about the intended message are expected to facilitate fast and effortless comprehension \citep{balota1994visual}.
Conversely, degraded visual signals---caused by factors such as lighting, occlusion, or visual interference---are likely to increase processing effort and raise the likelihood of errorful reading.

This prediction fits well within noisy channel models of reading.
In a noisy channel model \citep{shannon1948mathematical}, a message is encoded and sent over a channel, where it is potentially corrupted.
A receiver, at the other end of the channel, must decode the most probable intended message given the received inputs.
Previous work has looked at the role of noise during reading, demonstrating how noise over uncertain inputs can lead to non-veridical interpretations \citep{levy-2008-noisy, gibson2013channel}.

While intuitive, to the best of our knowledge, the impact of noisy inputs on reading effort has not been quantified within a formal computational model of reading.
That is, although many theories of reading assume that poorer sensory input leads to more effortful processing, and classic experimental work has shown that reduced visual signals increase processing difficulty and interact with other lexical properties \citep{rumelhart1974process}, they have not derived or tested this relationship quantitatively. 
In this paper, we aim to fill this gap by providing an information-theoretically grounded, quantitative account of how bottom-up input quality affects processing effort. 
Our central proposal is that input quality can be formalized as the mutual information (MI) between (visual) input and word identity. 
From an information-theoretic perspective, a signal is informative to the extent that it reduces uncertainty about a target variable---in this case, the identity of a word. 
We assume that greater processing effort manifests in longer reading times, and therefore predict that reductions in mutual information should lead to systematic slowdowns in reading.

This paper makes three contributions:
First, we instantiate the above operationalization of visual input quality in reading under a formal model of reading as a Bayesian update.
Second, we provide a quantitative estimate of the cost of reduced input quality on processing effort.
To do so, we use multimodal language models to estimate mutual information over a dataset of partially masked word images.
We then collect human reading times on the same stimuli, using the MoTR paradigm \citep{wilcox2024mouse}, which simulates eye-tracking, and can be used to collect data over the web.
We use these data to estimate the relationship as a specific slowdown in terms of nats of information gain (the pointwise variant of mutual information) per millisecond of processing time.
Our data suggest that the cost of reduced information is not linear---small losses in informational quality can lead to disproportionately large increases in reading time, particularly in the upper regions of a signal's informational range.

Our third contribution is to compare how information is distributed across visual forms of words in two typologically distinct languages.
To that end, we collect data in both English and Chinese, representing alphabetic and logographic scripts, respectively.
We find that, in both languages, the upper half of a word contains more information about word identity than the lower half.
However, the asymmetry is more pronounced in English than in Chinese, a pattern that is reflected in the reading times.

\section{Formal Model} \label{sec:formal_model}

\newcommand{\tstep}{\textcolor{\mathcolour}{\ensuremath{t}}\xspace}
\newcommand{\word}{\textcolor{\mathcolour}{\ensuremath{w}}\xspace}
\newcommand{\words}{\textcolor{\mathcolour}{\ensuremath{\mathbf{w}}}\xspace}
\newcommand{\prevwords}{\textcolor{\mathcolour}{\ensuremath{\mathbf{w}_{<\tstep}}}\xspace}
\newcommand{\wt}{\textcolor{\mathcolour}{\ensuremath{\word_{\tstep}}}\xspace}

\newcommand{\Word}{\textcolor{\mathcolour}{\ensuremath{W}}\xspace}
\newcommand{\Wt}{\textcolor{\mathcolour}{\ensuremath{\Word_{\tstep}}}\xspace}

\newcommand{\Words}{\textcolor{\mathcolour}{\ensuremath{\mathcal{W}}}\xspace}
\newcommand{\Evid}{\textcolor{\mathcolour}{\ensuremath{E}}\xspace}
\newcommand{\Evids}{\textcolor{\mathcolour}{\ensuremath{\mathbf{E}}}\xspace}
\newcommand{\evid}{\textcolor{\mathcolour}{\ensuremath{e}}\xspace}
\newcommand{\evids}{\textcolor{\mathcolour}{\ensuremath{\mathbf{e}}}\xspace}
\newcommand{\prevevid}{\textcolor{\mathcolour}{\ensuremath{\evids_{1:t-1}}}\xspace}
\newcommand{\nsamples}{\textcolor{\mathcolour}{\ensuremath{k}}\xspace}
\newcommand{\trueinput}{\textcolor{\mathcolour}{\ensuremath{w^*}}\xspace}
\newcommand{\E}{\textcolor{\mathcolour}{\ensuremath{\mathbb{E}}}\xspace}
\newcommand{\threshold}{\textcolor{\mathcolour}{\ensuremath{\phi}}\xspace}

\newcommand{\expect}[1]{\textcolor{\mathcolour}{\ensuremath{\mathbb{E}_{#1}}}\xspace}

\newcommand{\mi}{\textcolor{\mathcolour}{\ensuremath{I}}\xspace}

\subsection{Reading as Bayesian Update}\label{subsec:Bayesian}

Following an extensive prior literature \citep{norris2006bayesian, bicknell2010rational, gauthier-levy-2023-neural}, we model word recognition as a Bayesian update process. 
Readers incrementally process a word \word drawn from a vocabulary \Words, where $\word \in \Words$ denotes a realization of a random variable \Word taking values in \Words.
We refer to a word at a particular timestep, \tstep, as \wt and the corresponding random variable at this timestep as \Wt.
We assume that readers intake individual samples of input \evid, where $\evid \in \mathbb{R}$ denotes a realization of a random variable \Evid ranging over the samples\footnote{For simplicity, we model inputs as continuous and univariate. However, we acknowledge that inputs may be more aptly modeled as multivariate and see this as an easy extension of the formal presentation given here.}.
Input samples could be either a patch of visual input for visual reading or a haptic percept in the case of braille.
Following previous work (e.g., \citealp{bicknell2010rational}), we model the process of reading as one of sequential word identification given input \evid and a previous context of words \prevwords.
In such models, readers are assumed to rationally integrate their prior expectations about a word, $P(\wt \mid \prevwords)$, with the likelihood of the observed input \evid, $P(\evid \mid \wt, \prevwords)$.
Instead of a single sample, we assume that readers integrate evidence over $\nsamples$ samples, $\evids_{1:\nsamples}$.
The rational update process we use to model reading is therefore: 
\begin{align}
    P(\wt \mid \evids_{1:\nsamples}, \prevwords) & \propto \\
    P(\wt \mid \prevwords) & \times \prod_{i=1}^{\nsamples} P(\evid_i \mid \wt, \prevwords) \nonumber
\end{align}

This tells us how readers update beliefs about a word given inputs and priors.
But reading is a dynamic process. How do readers choose when to move on to the next word?
Previous work models this by proposing that readers draw samples until the uncertainty about the current word reaches a threshold, \threshold, at which point they move on (e.g., \citealp{li2024information}).
We quantify uncertainty as the entropy of the posterior distribution. 
That is, sampling continues until:
\begin{align}
    H(P(\word_{\tstep} \mid \evids_{1:\nsamples}, \prevwords)) \leq \threshold
\end{align}

However, given a particular actual input \trueinput we cannot be certain how many samples a reader draws or what information each sample contains. To account for this uncertainty, we therefore make the prediction that readers will move on when the \emph{expected} entropy falls below this threshold, where the expectation is taken over uncertain inputs:
\begin{align}
    \expect{\Evids_{1:\nsamples}}[H(\Wt \mid \Evids_{1:\nsamples}, \prevwords)] \leq \threshold
\end{align}

Although we assume that reading does take place given a context, for the rest of this section, we will drop the word-context term, \prevwords. We note that it would be easy to add this term back into the subsequent equations as a conditioning variable without changing the overall model.

\subsection{Quality of Bottom-Up Evidence}

We model the quality of the inputs as the mutual information between the inputs and the word identity, i.e., as $\mi(\Word;\Evid)$.
High-quality inputs do a better job of reducing uncertainty over words.
For a given word-identification step, we can write the mutual information between a word and the total number of samples drawn as $\mi(\Word;\Evids_{1:\nsamples})$.
Using the chain rule of mutual information \citep{cover1999elements} and assuming that there is \emph{conditional independence} between samples, given \Word, we can derive the following inequality:\footnote{For more discussion of these assumptions, see \Cref{app:assumptions}.}
\begin{subequations}
\begin{align}
    \mi(\Word ; \Evids_{1:\nsamples}) &= \sum_{i = 1}^{\nsamples} \mi(\Word ; \Evid_i \mid \Evids_{1:i-1}) \label{eq:mi_sum} \\
    \text{\footnotesize{\textcolor{gray}{assuming}}} \atop \text{\footnotesize{\textcolor{gray}{cond. independence}}}&\leq \sum_{i = 1}^{\nsamples} \mi(\Word ; \Evid_i) \label{eq:mi_cond_ind} \\
    &\leq \nsamples \times \mi(\Word ; \Evid) \label{eq:iid_mi};
\end{align}
\end{subequations}

How is the mutual information between inputs and words related to the reading process, as described above? 
We assume that taking samples and processing these samples takes cognitive effort.
Following previous work, we also assume a link between effort and time \citep{levy2008expectation, hale-2001-probabilistic}.
Therefore, the more samples, \nsamples, a reader needs to take in order to reduce uncertainty, the longer it will take them to read a given word.

We can now link the quality of inputs to our reading process through the definition of mutual information:
\begin{align}
    \mi(\Word ; \Evids_{1:\nsamples}) &= H(\Word) - H(\Word \mid \Evids_{1:\nsamples})
\end{align}

\noindent Plugging in the inequality from \ref{eq:iid_mi}, and the definition of conditional entropy,\footnote{That is: $H(X \mid Y) = \expect{Y}[H(X \mid Y)]$.} we rearrange the terms:
\begin{align}
    \expect{\Evids_{ 1:\nsamples}}[H(\Word \mid \Evids_{1:\nsamples})] \geq
    H(\Word) - \nsamples \times \mi(\Word ; \Evid)
\end{align}

That is, the expected entropy of the posterior distribution, given uncertain inputs, is greater than the entropy over words minus the number of samples taken times the mutual information between the samples and the words.

For our model of reading, we are interested in when the entropy of the posterior distribution is approximately \threshold.
In particular, we are interested in how many samples must be drawn to reach this threshold, as this determines the effort (and therefore the time) required to reduce uncertainty enough to move on to the subsequent word.
Substituting in our threshold parameter in and rearranging the terms, we have:
\begin{align}
    \nsamples \geq \frac{H(\Word) - \threshold}{\mi(\Word ; \Evid)}
\end{align}

The minimum number of samples required to reach the threshold grows with the entropy of the distribution over \Word. 
Likewise, it decreases with the mutual information between \Word and \Evid.
Because we assume a link among the number of samples, effort and time, this leads us to the following two predictions:

\begin{pred} \label{pred:entropy}
    \textbf{Top-Down Processing \& Entropy:} As the entropy of a word-position \Word increases, average reading time increases.
\end{pred}

\begin{pred} \label{pred:mi}
    \textbf{Bottom-up Processing \& Mutual Information:} As the mutual information between words \Word and their visual representations \Evid decreases, average reading time increases.
\end{pred}

In fact, \Cref{pred:entropy} has already been investigated by \citet{pimentel2023anticipation}, whose results confirm our prediction.
\citeauthor{pimentel2023anticipation} refer to the entropy over the next word, given a set of previous words $H(\Wt \mid \prevwords)$ as a word's \emph{contextual entropy}.
They find that as word-level contextual entropy increases, so too does reading time.
For the rest of this paper, therefore, we are interested in testing \Cref{pred:mi}, namely whether the quality of bottom-up evidence, modeled as mutual information between words and visual information, affects word-by-word reading times.

\section{Methods}

\subsection{Materials} \label{sec:methods}

We use a portion of the advanced OneStopQA dataset \citep{berzak-etal-2020-starc}.
This dataset contains Guardian news articles, along with carefully constructed reading comprehension questions, which are linked to individual spans in the text.
For our study, we selected three articles: “101-Year-Old Bottle Message”, “Inky the Octopus Escapes from Aquarium”, and “Japan Calls Time on Long Hours Work Culture”. A team member with experience in English-Chinese translation hand-translated these texts and their questions into Mandarin. 
This small translated corpus, which we term the \defn{Chinese OneStopQA}, is released along with the publication of this article (see code repository).

The English subset contains 1,793 words (mean word length $= 4.6$, $SD=2.53$), while the Chinese subset contains 3,182 characters. In terms of experimental presentation, one Chinese character occupies roughly 1.46 times the pixel space of an English letter, making an average English word about 3.2 times longer than a Chinese character. The average Zipf frequency is slightly higher in English ($M=5.77$, $SD=1.45$) than in Chinese ($M=4.84$, $SD=1.90$), largely due to the low frequency of transliterated Western named entities in the Chinese translations.

\begin{figure}
    \centering
    \includegraphics[width=0.95\columnwidth]{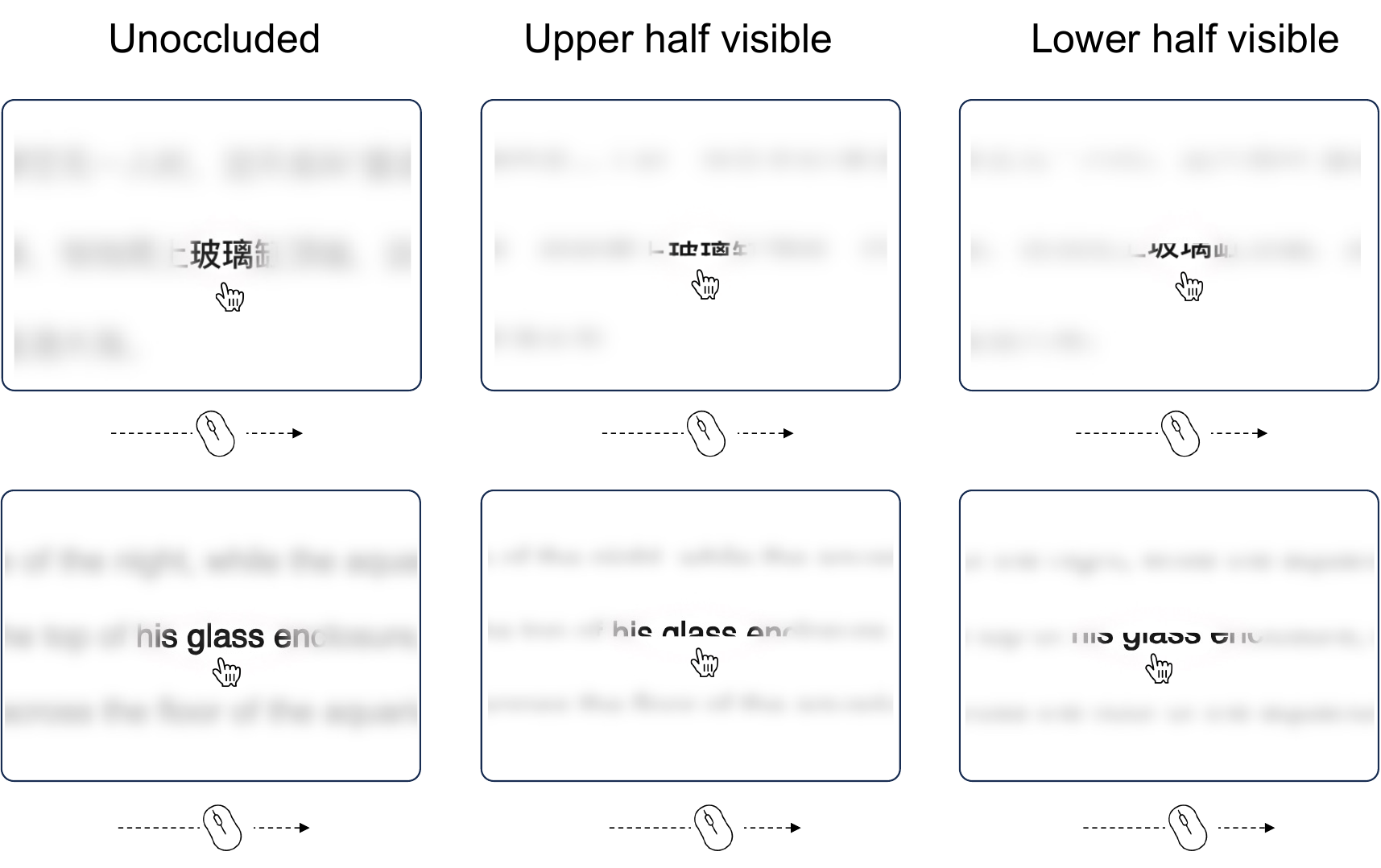}
    \caption{Example showing a screen from a MoTR trial with our three different reading conditions.}
    \label{fig:motr_conditions}
\end{figure}

\paragraph{Creating Noisy Words}

To create noised reading conditions, we occluded (i.e., masked with white) either the upper or lower half of every word in the dataset.
There are potentially many ways to add noise to the texts.
Other options would be to occlude the first half or the second half of words, as well as Gaussian noise.
Previous work has shown that the beginnings of words tend to carry more disambiguatory information than their endings. For example, \citet{pimentel-etal-2021-disambiguatory} demonstrates this cross-linguistically using information-theoretic measures, while \citet{alhama2019role} presents a perceptually constrained connectionist model explaining fixation biases toward word onsets. These findings are consistent with psycholinguistic evidence of the optimal viewing position effect in visual word recognition\citep{brysbaert2005visual}. \looseness=-1

However, these studies focus on the linear distribution of information across letter positions, which applies naturally to alphabetic scripts such as English but not to logographic systems like Chinese, where characters are two-dimensional and not arranged linearly. We were also concerned that completely removing some letters or characters would make reading too difficult or frustrating for participants, and that the removal of letters or characters demands very careful handling to avoid confounds \citep{rayner1998eye, rayner2006raeding}.
Masking the upper or lower half retains some information about each character, which presents a paradigm that is relatively readable for participants, especially in the degraded conditions. In addition, unlike simply adding Gaussian noise, upper and lower half occlusion allows us to investigate \emph{where}, in vertical space, information is localized in English and Chinese orthographic systems.
Our strategy leads to two additional research questions:
\begin{subresearchq}
    Is information split up differentially between the upper and lower halves of orthographic words?
\end{subresearchq}
\begin{subresearchq}
    Does the relative informativeness of upper vs. lower halves differ across languages?
\end{subresearchq}

\subsection{Data Collection}

\paragraph{Mouse Tracking for Reading (MoTR)}

To test our main predictions, we need a way of measuring (average) human reading times in our different conditions.
To do so, we use Mouse Tracking for Reading \citep[MoTR;][]{wilcox2024mouse}.
In a MoTR trial, a blurred text is presented on a screen.
A small region around the tip of a user's mouse brings the text into focus.
Participants move the mouse to incrementally \emph{reveal} and read the text, while their mouse location is recorded and used as a proxy for gaze location.
The time-stamped $x/y$ coordinates are then 
turned into incremental word-by-word reading times, similar to word-level reading times in an eye-tracking-while-reading experiment. 
As in eye-tracking, there are several ways to compute reading times.
For our main analysis, we use the \defn{first-pass reveal time (FPRT)}, defined as the total amount of time a participant spends revealing a word during their first pass reading. Conveniently, the same acronym (FPRT) is used in eye-tracking for \defn{first-pass reading time}, but we use ``reveal'' to emphasize that it is computed from mouse movements rather than eye fixations.


\citet{wilcox2024mouse} show that MoTR reading measures are strongly correlated with eye-tracking and self-paced-reading measures.
MoTR has been used to collect data in English and Russian \citep{ouguz2025using}, but not in Chinese.

\paragraph{Participants}

We recruited $54$ English and $57$ Chinese speakers on Prolific, requiring a minimum approval rate of 98\% and the corresponding language to be their first and native language. Participants were compensated 3.75 GBP for a median reading time of 25 minutes. 

\paragraph{Procedure}
Each participant read the article paragraphs presented screen by screen, with each screen randomly assigned to one of three conditions: upper-half occluded (i.e., lower-half visible), lower-half occluded (i.e., upper-half visible), or unoccluded (see Figure~\ref{fig:motr_conditions}). In addition to reading texts and answering comprehension questions, we ask participants to rate the ease of reading after finishing all the trials using a multiple-choice question: ``Which do you find easier to read: text showing only the top half or only the bottom half?'' The options are ``Top half only,'' ``Bottom half only,'' and ``About the same.'' The actual experiments are available online for Chinese\footnote{https://cuierd.github.io/Re-Veil/multilingual\_motr/zh/} 
and English\footnote{https://cuierd.github.io/Re-Veil/multilingual\_motr/en/}.

\subsection{Mutual Information Estimation}

\newcommand{\Orth}{\textcolor{\mathcolour}{\ensuremath{\mathbf{O}}}\xspace}
\newcommand{\orth}{\textcolor{\mathcolour}{\ensuremath{\mathbf{o}}}\xspace}
\newcommand{\params}{\textcolor{\mathcolour}{\ensuremath{\mathbf{\theta}}}\xspace}
\newcommand{\surp}{\textcolor{\mathcolour}{\ensuremath{\iota}}\xspace}

In \Cref{sec:formal_model}, our model concerns the mutual information between words, \Word, and (visual) evidence sampled by the reader, \Evids.
However, we do not have direct access to this evidence.
Instead, as a proxy for our visual evidence, we estimate the mutual information between words \Word and their orthographic representations $\orth \in \mathbb{R}^d$, where \orth is a realization of a random variable \Orth that ranges over representations of different words. 
Following \citet{pimentel-etal-2020-information}, we decompose the mutual information as
\begin{subequations}
    \begin{align}
    \mi(\Word;\Orth) &= H(\Word) - H(\Word \mid \Orth) \\
    & \approx H_{\params}(\Word) - H_{\params}(\Word \mid \Orth)
\end{align}
\end{subequations}

\noindent where \params denotes the parameters of the models employed for entropy estimation. We estimate each of these two terms separately.

We estimate \textbf{unconditional entropy} $H_{\params}(\Word)$ with a maximum likelihood estimation of the unigram distribution of Chinese characters and English words. We take the $9,933$ unique Chinese characters included in the modern Chinese character database\footnote{https://lingua.mtsu.edu/chinese-computing/}, and the $60,384$ English words in the SUBTLEXus database \citep{Brysbaert2024the}, and look up their frequencies using the Python library \textit{wordfreq} \citep{robyn_speer_2022_7199437} that supports both languages and aggregates data from multiple domains, including subtitles, Wikipedia, news, fiction, and web content. 
Normalizing the frequencies, we obtain the empirical distribution $p_{\theta}(\word)$ and from it we can directly compute the entropy $H_{\params}(\Word)$. The empirical entropies are $5.59$ and $7.12$ nats for Chinese characters and English words.

We estimate the \textbf{conditional entropy} $H_{\params}(\Word \mid \Orth)$ in two stages.
First, we compute the word entropy conditioned on a specific orthographic representation, $H_{\params}(\Word 
\mid \Orth=\orth)$. We refer to this as \defn{pointwise conditional entropy}. We compute this value by taking the expectation of the information content, or \defn{surprisal} of all words given the orthographic representation $\surp_{\params}(\word \mid \orth)$, where $\surp_{\params}(\cdot) = - \log p_{\params}(\cdot)$.  
Given a model with parameters \params that can produce our probability distribution of interest, that is, $p_{\params}(\word \mid \orth)$, the pointwise conditional entropy is calculated as:
\begin{align}
    &H_{\params}(\Word \mid \orth) \approx \sum_{\word \in \Words} p_{\params}(\word \mid \orth)\surp_{\params}(\word \mid \orth)
    \label{eq:unconditional_entropy}
\end{align}

We then estimate conditional entropy as the expectation of the pointwise conditional entropy with respect to \Orth, following the identity $H(\Word \mid \Orth)=\expect{\Orth}[H(\Word \mid \Orth = \orth)]$.
We take the expectation over a set of held-out test samples:
\begin{align}
    &H_{\params}(\Word \mid \Orth) \approx \frac{1}{N} \sum_{n=1}^{N} H_{\params}(\Word \mid \orth^n)
\end{align}

\noindent where $\orth^n$ is the $n^{th}$ orthographic representation in the test set. 

We note that using these methods, we can estimate not only the mutual information $\mi(\Word;\Orth)$, but also its pointwise variant, also called the \defn{information gain (IG)}, for a particular orthographic representation, where $IG(\Word;\orth) = H(\Word) - H(\Word \mid \orth)$. While our formal prediction is made in terms of mutual information, in \Cref{sec:effect_result}, we use IG to investigate the relationship between information contained in individual visual inputs and their respective reading times. \looseness=-1

In recent work, similar methods have been used to study the relationship between words (as represented by text) and prosody, or the melody of speech \citep{wolf-etal-2023-quantifying, regev2025time, wilcox2025using}.
However, these previous works learn distributions over real-valued variables that represent pitch.
We wish to learn distributions over discrete \word-valued variables $p_{\params}(\word \mid \orth)$.
To obtain this distribution, we use multimodal language models, which we fine-tune to produce conditionalized distributions over words, given visual inputs. We do so with the following methods:

\begin{figure}
    \centering
    \includegraphics[width=0.99\columnwidth]{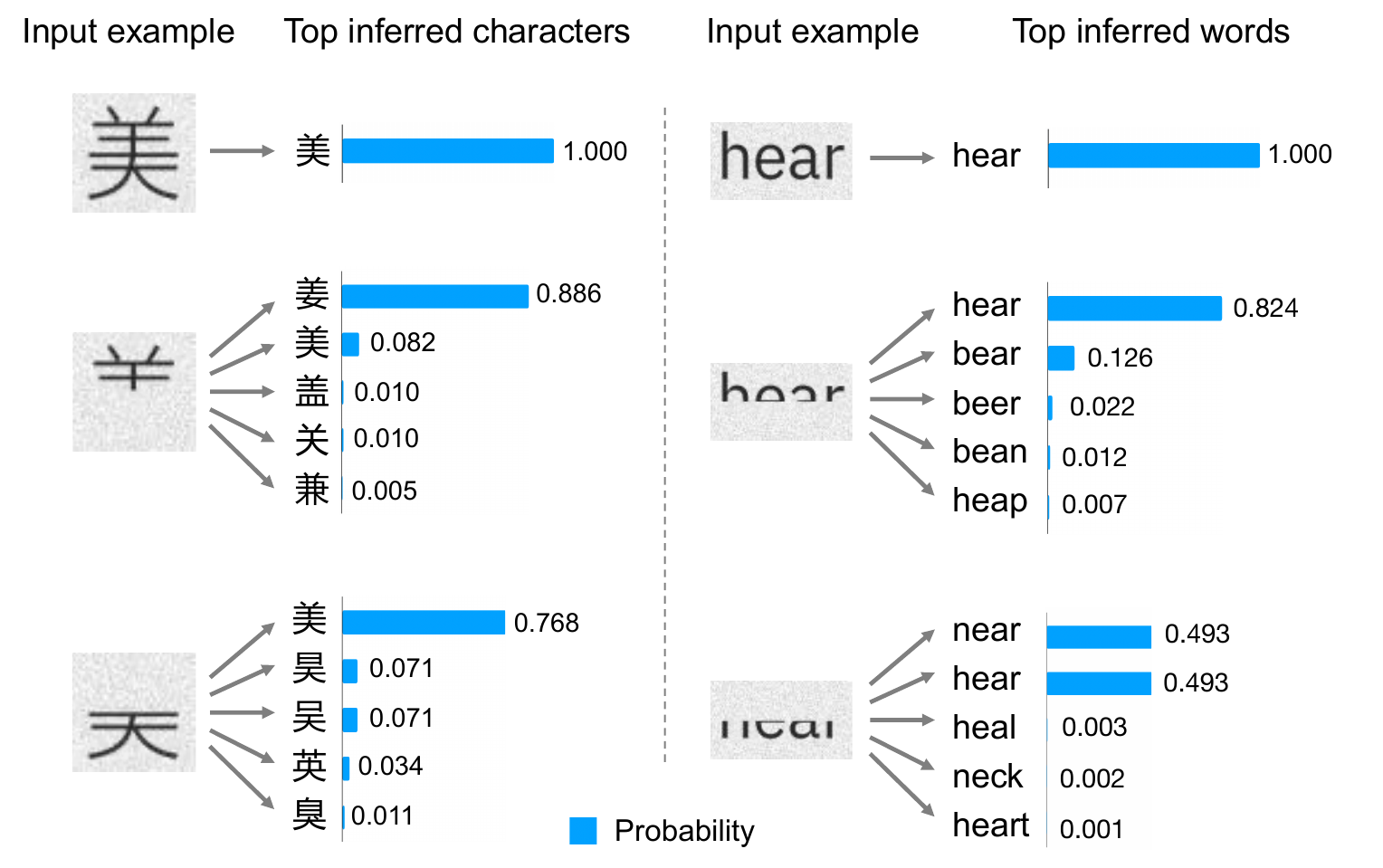}
    \caption{Results of fine-tuned Qwen2.5 model for  the Chinese character \begin{CJK}{UTF8}{gbsn}美\end{CJK} (``beautiful'') and the English word \emph{hear}. The preference for \emph{hear} over \emph{heal} in upper half occlusion likely reflects pre-training frequency bias, which we control for by training TransOCR from scratch.}
    \label{fig:Qwen_inference_example}
\end{figure}

\paragraph{Data Generation}
We adapt the Python library \textit{TRDG}\footnote{https://github.com/Belval/TextRecognitionDataGenerator} to generate images of Chinese characters and English words from text as their orthographic representations, applying upper-, lower-half occlusion to create our different experimental conditions. For each character or word under each condition, we generated six images with different fonts and font weights\footnote{For Chinese, the fonts are XinYiJiXiangSong, FZHei-B01, FZKai-Z03, NotoSansSC-Regular, NotoSerifSC-SemiBold, and SourceHanSans. For English, they are DroidSans, Lato-Bold, NotoSans, PTSerif, Raleway, and Sansation.} to enhance visual variability, and added a small amount of Gaussian noise to the image backgrounds \citep{li2025texture}.
We generate $16,800$ Chinese character images and $44,800$ English word images under each of the three occlusion conditions as training data. For test data, we generate images of all Chinese characters in the Chinese OneStopQA dataset and all English words in the selected OneStopQA subset, again under each occlusion condition.

\paragraph{A Bayesian Baseline Model}
As a simple reference, we implement a Bayesian baseline to estimate the pointwise conditional entropy $H(\Word \mid \Orth=\orth)$. In this model, lexical frequencies provide priors $p(\word)$, and structural similarity \citep[SSIM;][]{wang2004image}, computed with \textit{scikit-image} \citep{van2014scikit}, serves as a likelihood $p(\orth \mid \word)$. Posterior probabilities $p(\word \mid \orth)$ are obtained by normalizing the product of priors and likelihoods across all candidate images, with the denominator corresponding to the marginal likelihood $p(\orth)$. We then compute $H(\Word \mid \Orth=\orth)$ from these posteriors for each input image and average across the dataset to obtain $H(\Word \mid \Orth)$. While straightforward, this baseline has some limitations. First, its computational cost scales quadratically with dataset size (O(N²)). Second, SSIM captures only low-level pixel similarity, often producing clustered scores for orthographically distinct characters or words. Third, its estimates are largely driven by lexical frequency. For these reasons, we turn to multimodal models based on artificial neural networks (ANNs) for more reliable and scalable estimation.

\paragraph{ANN-based Predictive Multimodal Models}
We use three different multimodal model settings:
First, we evaluate the pre-trained Qwen2.5-VL-7B-Instruct\footnote{https://huggingface.co/Qwen/Qwen2.5-VL-7B-Instruct} in a zero-shot setting. Qwen2.5-VL-7B is an open-source vision-language model developed by Alibaba, designed for high-accuracy multimodal analysis with enhanced visual understanding and text-image alignment \citep{wang2024qwen2, bai2025qwen2}.
As upper- and lower-half occluded words are likely out-of-distribution with respect to the model's training data, we do not expect the mutual information estimate to be tight in this setting.
For a better estimate, we then fine-tune Qwen2.5-VL-7B on our task-specific data to improve its performance. 
To complement the estimate from the pre-trained model,  we also train a separate transformer-based OCR model \citep[TransOCR;][]{Yu2023scene}, from scratch, to perform the same prediction task. The model combines a ResNet encoder with a Transformer decoder for character recognition. 
Full training configurations and prompt designs for the Qwen and TransOCR models are provided in \Cref{app:fine_tuning} and \Cref{app:ocr_training}, respectively.


\section{Results}

\subsection{Human Reading Results}

\begin{figure*}
    \centering
    \includegraphics[width=0.95\linewidth]{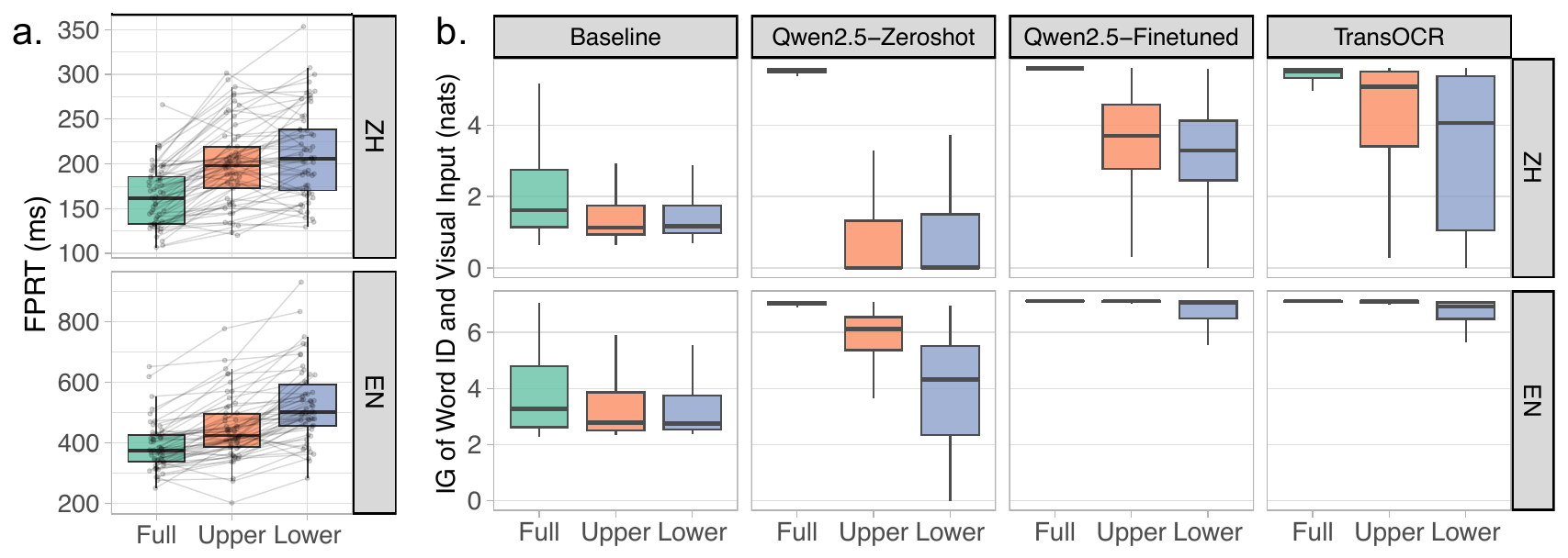}
    \caption{\textbf{(a)} Reading times (FPRT) measured under three visibility conditions. Boxes represent the interquartile range (middle 50\%), center lines indicate the median, and whiskers show the overall data spread. Grey lines trace each participant’s mean across conditions. EN: English; ZH: Simplified Chinese \textbf{(b)} Information gain (IG) between word identity and visual form under the three conditions, obtained with each of our estimation techniques.}
    \label{fig:human_and_model}
\end{figure*}

We show reading times in Figure~\ref{fig:human_and_model}(a). In both languages, reading full words resulted in the shortest average FPRT, as predicted. Interestingly, both languages follow a \emph{Full} < \emph{Upper} < \emph{Lower} pattern, with lower-half visibility leading to the longest FPRT. To quantify these effects, we fit linear mixed-effects models with visibility condition as a predictor, using sliding contrasts to compare \emph{Upper} vs. \emph{Full} and \emph{Lower} vs. \emph{Upper}. Word length (EN only), lexical frequency, surprisal, and contextual entropy are included as control variables, with random intercepts for subjects and items. In Chinese, both contrasts are significant: $\beta = 37.79$\,ms ($p < 2e{-16}$) and $\beta = 12.64$\,ms ($p < 2e{-16}$). In English, the effects are larger: $\beta = 69.93$\,ms ($p < 2e{-16}$) and $\beta = 90.09$\,ms ($p < 2e{-16}$)\footnote{FPRT is calculated for Chinese \emph{characters} and English \emph{words}, which may explain the generally longer reading times in English.}.

These results can be interpreted as implying a visual asymmetry in both languages between ease of processing with respect to just upper and lower halves of words. The asymmetry is stronger in English, where the lower half leads to greater slowdowns. Participants' subjective ratings confirm this asymmetric pattern and further show that English lower halves are perceived as harder to read than Chinese ones (Appendix~\ref{app:subjective_ease}).
In addition, we summarize comprehension question performance in Appendix~\ref{app:RCQ_performace_table}. Accuracy declines with occlusion, but remains well above chance (25\%), indicating that reading is effective.

\subsection{Mutual Information Results}

\begin{table}[ht]
\centering
\scriptsize
\setlength{\tabcolsep}{3pt} 
\begin{tabular}{cc|ccc|ccc}
\toprule
 & & \multicolumn{3}{c|}{Acc (\%)} & \multicolumn{3}{c}{MI (nat)} \\
\cmidrule(lr){3-5} \cmidrule(lr){6-8}
Language & Model & Full & Upper & Lower & Full & Upper & Lower \\
\midrule
 & Baseline 
   & - &  - & - & 4.85 & 4.42 & 4.41 \\
 & Qwen2.5-zs 
   & 97.9 & 5.2  & 3.2 & 5.42 & 0.27 & 0.32 \\
ZH & Qwen2.5-ft 
   & 99.0 & 48.6 & 44.3 & 5.57 & 3.62 & 3.27 \\
 & TransOCR 
   & 97.9 & 78.8 & 65.7 & 5.26 & 4.09 & 3.17 \\
\midrule
 & Baseline 
   & - &  - & - & 6.43 & 6.12 & 5.99 \\
 & Qwen2.5-zs 
   & 98.8 & 84.1 & 51.5 & 6.99 & 5.74 & 3.86 \\
EN & Qwen2.5-ft 
   & 99.7 & 93.1 & 68.1 & 7.11 & 7.01 & 6.66 \\
 & TransOCR 
   & 98.8 & 95.7 & 65.8 & 7.07 & 7.00 & 6.68 \\
\bottomrule
\end{tabular}
\caption{Model accuracy (\%) and mutual information estimates, $I(\Word;\Orth)$, for Chinese (ZH) and English (EN) with different models.}
\label{tab:mi_combined}
\end{table}

To give a visual sense of how our models perform, in \Cref{fig:Qwen_inference_example}, we show sample images in the three experimental conditions, along with the predictions from the fine-tuned Qwen2.5 model. As a performance check, we also report the model accuracies (Acc) in \Cref{tab:mi_combined}. Baseline accuracies are not reported because they only compare each test image against other images in the test set, rather than predicting over the full vocabulary. This makes its accuracy values not directly comparable to the other models. Accuracies in the unoccluded Full condition are uniformly high (>97\%). Under occlusion, zero-shot Qwen2.5 drops dramatically in Chinese (<6\%) but is still strong in English (52--85\%). Fine-tuning improves performance in both languages, while TransOCR achieves the most robust accuracy overall.

Our main analyses focus on mutual information, \mi(\Word;\Orth) in \Cref{tab:mi_combined} and information gain, $IG(\Word;\orth)$ in Figure~\ref{fig:human_and_model}(b), across conditions and models. Figure~\ref{fig:human_and_model}(b) shows IG between word identity and visual input under three visibility conditions, estimated with the Bayesian baseline, Qwen2.5-VL-7B-Instruct (zero-shot and fine-tuned), and TransOCR. \Cref{tab:mi_combined} reports MI as the IG averaged across all inputs. MI decreases systematically from \emph{Full} to \emph{Upper} to \emph{Lower} in both languages.  

\begin{figure*}[!t]
    \centering
    \includegraphics[width=0.95\linewidth]{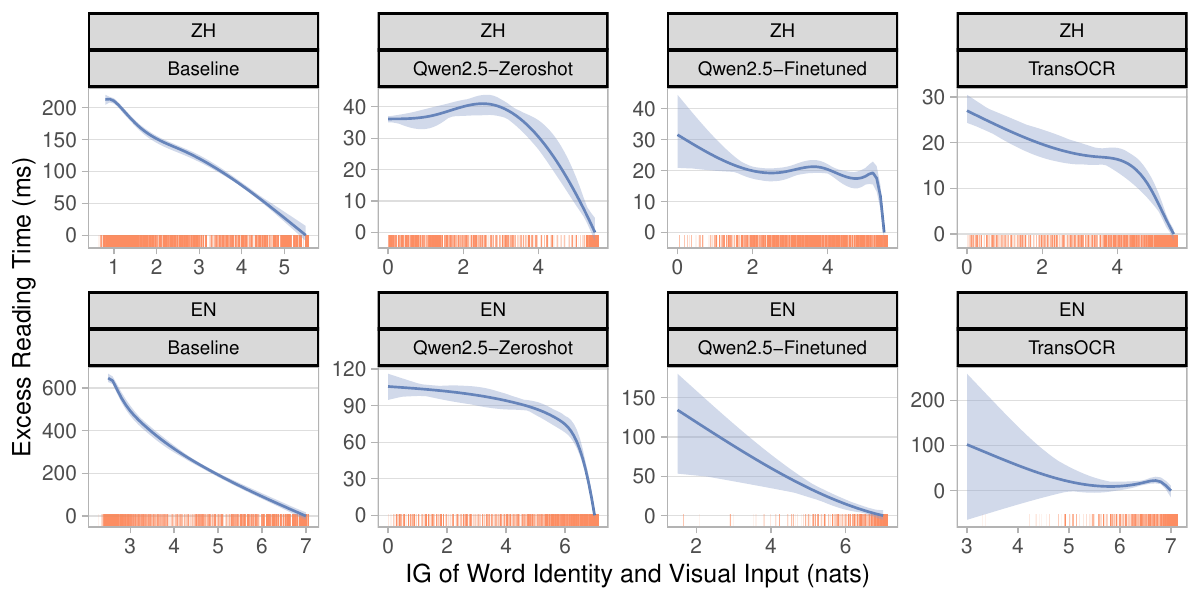}
    \caption{Relationship between informational quality of individual words (information gain; IG) and excess reading time.
    Solid blue lines are smoothed GAM fits; shaded regions show bootstrapped 95\% confidence intervals. 
    Red tick marks along the bottom (rug plots) indicate the distribution of IG data points.
    Reading times are aligned to end at zero at the highest MI end to emphasize the relative excess reading time when information quality decreases.
    }
    \label{fig:rt_visual_mi_link}
\end{figure*}

For a statistical test of our observed trends, we fit linear mixed-effects models for each language--model pair, with visibility as the main predictor. As in the reading time analysis, we use the same sliding contrasts. Word length (EN only), lexical frequency, surprisal, and contextual entropy are included as controls, with random intercepts for items.\footnote{Empirically, IG shows little correlation with these controls ($-.03$ to $.05$ in Chinese; $-.22$ to $.15$ in English). An exception is the Bayesian baseline, where IG correlates strongly with frequency (>0.9) in both languages.}
In Chinese, all models show significant IG reductions when only the upper half is visible (Baseline: $\beta = -.58$; Qwen2.5-Zeroshot: $\beta = -4.55$; Qwen2.5-Finetuned: $\beta = -1.85$; TransOCR: $\beta = -.99$ nats), and IG from fine-tuned models dropped further when visibility changed from Upper to Lower (Qwen2.5-Finetuned: $\beta = -.37$; TransOCR: $\beta = -1.01$ nats). In English, the zero-shot model showed the largest overall drop (Upper vs. Full: $\beta = -1.43$; Lower vs. Upper: $\beta = -2.09$ nats), while the baseline and fine-tuned models show smaller but consistent reductions (Baseline: $\beta = -.36$, $-.07$; Qwen2.5-Finetuned: $\beta = -.11$, $-.47$; TransOCR: $\beta = -.08$, $-.35$ nats).
All effects were statistically significant at $p < .0001$.
Panels (a) and (b) of Figure~\ref{fig:human_and_model}, taken together, reveal a clear pattern: as visual input degrades from \emph{Full} to \emph{Upper} to \emph{Lower}, as measured by IG, reading times increase.

\subsection{Word-Level Relationship} \label{sec:effect_result}

In this section, we test the relationship between reading time and informational quality at the \emph{word level}.
To do so, we fit linear mixed-effects models with reading time of an orthographic representation as the dependent variable and its IG as a fixed effect. We also included frequency, surprisal, contextual entropy, and (in EN) word length as additional fixed effects, as well as by-subject and by-item random intercepts.

We find a significant effect of IG on reading time across all models and measures, with a consistent negative effect: the higher the informational quality of the input, the faster it is read.
In Chinese, all four IG estimates were significant predictors of FPRT: $\beta = -14.94$ ms (Baseline), $\beta = -7.53$ ms (Qwen2.5-Zeroshot), $\beta = -10.19$ ms (Qwen2.5-Finetuned), and $\beta = -4.97$ ms (TransOCR). In English, the effects were even larger: $\beta = -15.36$ ms (Baseline), $\beta = -23.67$ ms (Qwen2.5-Zeroshot), $\beta = -51.48$ ms (Qwen2.5-Finetuned), and $\beta = -66.42$ ms (TransOCR). All effects were statistically significant at $p < .0001$.

\subsection{Nonlinear Relationship Between Information Quality and Reading Time}

While our linear regression models show that informational quality affects reading time, it makes strong assumptions about the functional form of this relationship.
In order to get a better sense of how these two variables are related, we visualize them together in Figure~\ref{fig:rt_visual_mi_link}.
We used generalized additive models (GAMs). GAMs are models that allow for non-linear relationships between predictor and response variables. We fit GAMs to predict reading times with smooth terms for IG, controlling for frequency, surprisal, contextual entropy, and (for English) word length. \footnote{For example, for EN data, the GAM was specified as: \texttt{FPRT \textasciitilde\ s(IG, bs=`cr', k=6) + te(freq, len, bs=`cr') + te(surp, len, bs=`cr') + te(entropy, len, bs=`cr')}.} 
We applied bootstrap smoothing over 20 resamples and computed confidence intervals for the estimated effects. 
We observe a consistent trend across both languages and all multimodal models: reading time remains relatively stable at lower IG estimates but decreases rapidly as IG increases in the upper end of its range. The Bayesian baseline does not show this pattern, as its IG values largely reflect lexical frequency.

\section{Discussion}

Turning back to our main prediction, we argue that our results provide converging evidence that visual quality, as measured by mutual information or information gain, impacts ease of processing.
First, we find a consistent ordering, both in terms of reading times and mutual information, across our three experimental conditions.
Second, we find a significant effect of the pointwise mutual information, or information gain, of individual words on reading times.
While intuitive, the idea that bottom-up informational quality impacts ease of reading has not been quantified within a formal framework of reading.
Our methods and experiments provide a specific estimate for the relationship between visual informational quality and reading times, which in English is between 25--66 ms/nat and in Chinese 5--10 ms/nat.
However, these numbers should be taken only as rough estimates, as the exact functional form may not be linear.

Turning now to our two sub research questions outlined in section \ref{sec:methods}: Interestingly, we find that information is not distributed evenly between the top and bottom halves of words.
Both English and Chinese place more information about word identity in the top half of their orthographic systems, a feature which we argue is reflected in the quicker reading times for our \emph{Upper} condition. This asymmetry may connect to more general visual biases, such as the top-heavy bias in object recognition \citep{viola2004can} or the upper visual field advantage \citep{previc1990functional}, as well as the way writing systems are shaped by reading and writing direction.

Interestingly, \citet{pimentel-etal-2021-disambiguatory} find similar informational asymmetries between the beginnings and ends of words, using an even wider set of languages.
Exploring whether their asymmetry in reading times and extending our results to more languages is an important direction for future research.
Finally, we find some suggestive evidence that this asymmetry is stronger in English, reflected in the larger effect sizes for the \emph{Upper} vs. \emph{Lower} contrast in our reading data. Future work should investigate such differences in greater detail.

\section*{Limitations}

There are several limitations with the present work. In our formal model, we made several assumptions: that visual samples of a given word \Evid are drawn i.i.d. during reading, and that visual inputs are conditionally independent from each other given \Word. While these assumptions are strong, they are compatible with a ``simple but fast'' approach to reading. We discuss them in more detail in \Cref{app:assumptions}. Moreover, our model assumes that the entropy threshold \threshold (Eq. 2) is always eventually reached, whereas in reality, readers may sometimes move on without reaching this threshold, especially under poor input quality. Besides, our model treats reading as a linear process, abstracting away from regressions, skips, and parafoveal preview. By focusing on FPRT as an index of early lexical processing, we capture word-level difficulty while necessarily ignoring the full dynamics of scanpaths. These simplifications keep the model straightforward but also mark directions for future work.

Another limitation concerns our approach to estimating mutual information between word identity and orthographic representation in Chinese.  We used characters, rather than lexical words, as the unit of analysis. This choice was motivated by two considerations: first, the average word length in our OneStopQA Chinese dataset is approximately 1.4 characters; second, Chinese characters, unlike English letters, carry substantial visual and semantic complexity. As such, characters may serve as a more suitable unit for modeling bottom-up visual processing in Chinese, analogous to words in English. Nonetheless, using lexical words might produce slightly different estimates of mutual information. Future work could examine whether similar patterns hold when words are used instead of characters.

Another limitation of the present work is that we did not extensively explore how top-down (contextual) processing can be integrated into the investigation of bottom-up processing. While much current research in psycholinguistics emphasizes the role of top-down expectations, our study is intended as a contribution specifically to the modeling of bottom-up processing. As described in Section \ref{subsec:Bayesian}, our formal Bayesian model is capable of incorporating contextual terms in a straightforward manner. However, for the scope of this analysis, we opted to exclude this component, leaving its exploration for future work.

Moreover, our study is limited to English and Chinese, as they are chosen to represent alphabetic and logographic systems, given the available corpora and our expertise. While this provides meaningful cross-linguistic insight, extending to more languages is important to generalize the conclusions of this work. In particular, the omission of Semitic languages is a notable gap, as their nonconcatenative morphology and distinct reading behaviors provide a critical test of generalizability (see e.g., \citealp{alhama2019role} and \citealp{lerner2014can}). Including such languages remains an important direction for future work.

\section*{Acknowledgments}
We thank the reviewers from ACL Rolling Review for their valuable feedback. We are also grateful to Kirill Semenov, Junlin Li, and our colleagues in the Computational Linguistics department at the University of Z\"urich for their helpful discussions and feedback. CD was supported by the MeRID grant (212276, PI: LJ).

\section*{Ethics Statement}

We do not foresee any specific ethical concerns deriving from this work. Experimental participants engaged in standard reading tasks, and were compensated for their time. We used AI models for analysis. As with all AI systems, it is important to acknowledge a general risk of bias or misuse.

\bibliography{acl_latex}

\appendix

\section{Assumptions of Formal Model} \label{app:assumptions}

\newcommand{\Ch}{\textcolor{\mathcolour}{\ensuremath{\mathbf{C}}}\xspace}
\newcommand{\ch}{\textcolor{\mathcolour}{\ensuremath{\mathbf{c}}}\xspace}

In this appendix, we discuss the assumption(s) of our formal model, namely that our samples \Evids are conditionally independent of each other, given \Word. First, we walk through the step from \ref{eq:mi_sum} to \ref{eq:mi_cond_ind}. We have by the definition of mutual information:
\begin{flalign}
& \mi(\Word; \Evids_{1:i}) \\
&= \sum_{i=1}^{\nsamples} \mi(\Word;\Evid_i \mid \Evids_{1:i-1})  \\
&= \sum_{i=1}^{\nsamples} H(\Evid_i\mid \Evids_{1:i-1}) - H(\Evid_i \mid \Word, \Evids_{1:i-1})
\end{flalign}

\noindent First the first term in this sum we can use the inequality $H(\Evid \mid \Evids_{1:i-1}) \leq H(\Evid)$. This is because adding information can only decrease entropy. Furthermore, assuming conditional independence between the samples, given \Word, we have that $H(\Evid \mid \Word, \Evids_{1:i-1}) = H(\Evid \mid \Word)$. Therefore, we can rewrite as:
\begin{flalign}
    &\leq \sum_{i=1}^{\nsamples} H(\Evid_i) - H(\Evid_i \mid \Word) \\
    &\leq \sum_{i=1}^{\nsamples} \mi(\Evid_i;\Word)
\end{flalign}

\noindent which, given the symmetry of mutual information, is what we have in \ref{eq:mi_cond_ind}.\footnote{Thank you to Tiago Pimentel for pointing out an earlier issue with our derivation, which has been corrected in this version of the paper. Any remaining mistakes are, of course, the fault of the authors.}

Regarding the assumption of conditional independence: This means that if the reader knows the word's identity, then previous samples do not necessarily predict what will be sampled next. We believe that this assumption is somewhat strong. However, it may be compatible with the view that readers adopt a simple, but fast, sampling strategy, in which prior evidence or even incremental knowledge about the word's identity from samples does not determine future sampling behavior. Given that reading happens at a very quick timescale, where word identification takes potentially only tens of milliseconds, such a ``simple but fast'' approach is not unreasonable.

\section{Qwen2.5-VL-7B-Instruct Fine-Tuning Details} \label{app:fine_tuning}
We fine-tune Qwen2.5-VL-7B-Instruct using QLoRA with 4-bit quantization and LoRA adapters applied to attention projection layers with rank 8, $\alpha=16$, and dropout 0.05. The model is trained for up to 100 epochs. Early stopping is applied based on validation loss. The training will terminate if no improvement for three consecutive epochs. AdamW (learning rate 2e-4), batch size 4, gradient accumulation of 8, and gradient clipping of 1.0.
Training data consists of system and user prompts with bottom-half character images; the model predicts a single Chinese character. We format inputs using Qwen’s chat template and pass them, along with images, through the model. The LM head outputs token-level logits, which are converted to probabilities. We compute cross-entropy loss directly on the gold assistant tokens, i.e., on the predictive distribution of the LM head, rather than on sampled outputs. Image inputs are processed using the Qwen processor. Training is conducted on a single GPU (RTX 3090 Ti).
Each training sample consists of a fixed system prompt and a task-specific user prompt. For example, for the lower-half recognition task, the templates used are as follows:
\paragraph{Chinese prompt}
\begin{quote}
\textbf{<system prompt>}
\begin{CJK}{UTF8}{gbsn}
你是一个善于识别汉字的智能助手。图片只展示了一个汉字的下半部分，请你根据下半部分准确识别该汉字，只回答一个汉字。
\end{CJK}

\textbf{<user prompt>}
\begin{CJK}{UTF8}{gbsn}
这张图片显示的是一个汉字的下半部分，上半部分被遮挡住了。请根据可见部分判断这是什么汉字，只回答一个汉字，不要包含其他内容。这个汉字是：
\end{CJK}
\end{quote}

\paragraph{English prompt}
\begin{quote}
\textbf{<system prompt>}
You are a helpful assistant that can identify English words in images. The image will show only the lower half of an English word, with the upper half masked. Identify the word accurately based on the visible portion. Please answer with a single word, and do not include any other text.

\textbf{<user prompt>}
The image contains the lower half of an English word. The upper half is masked. What is the word in the image? Please answer with a single word, and do not include any other text. The word is:
\end{quote}

\section{TransOCR Training Details} \label{app:ocr_training}

We trained the Transformer-based OCR model (TransOCR) for character recognition using the PyTorch framework. The model takes grayscale images resized to 32×256 pixels as input and is trained to predict character sequences in an autoregressive manner. Training was conducted using the Adadelta optimizer ($\rho$ = 0.9, weight decay = 1e-4) with an initial learning rate of 1.0 and a batch size of 16. The loss function was standard cross-entropy over predicted character classes. We applied early stopping with a patience of 5 epochs based on validation accuracy. 

All models were trained on two NVIDIA GPUs (RTX 3090 Ti) with multi-GPU support (DataParallel), and model checkpoints were saved at each epoch. The best-performing model was selected based on validation accuracy. 

During inference, character predictions were generated step-by-step. At each step, the model outputs a probability distribution over the character vocabulary via a softmax layer.  We denote this distribution given the image input \orth as $p_\theta(\ch\mid\orth)$, where \ch is a realization of a random variable \Ch ranging over characters. We first compute the character-level conditional entropy $H_\theta(\Ch \mid \orth)$ at each step using $H_\theta(\Ch\mid\orth) = - \sum_{\ch} p_\theta(\ch\mid\orth) \log p_\theta(\ch\mid\orth)$, and then sum up the entropies of all steps to obtain the word-level conditional entropy $H_\theta(\Word\mid\orth) = - \sum_{\Ch} H_\theta(\Ch\mid\orth)$.

\section{Self-Rated Ease of Reading} \label{app:subjective_ease}

\begin{figure}
    \centering
    \includegraphics[width=0.95\linewidth]{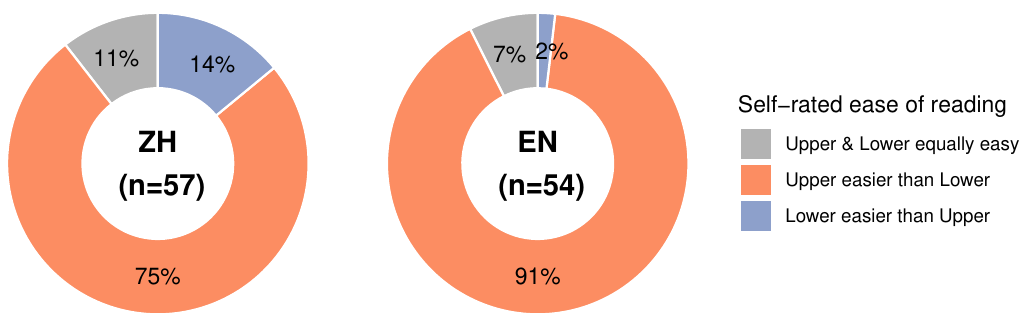}
    \caption{
    Self-rated ease of reading across visibility conditions. Participants were asked to judge whether the upper or lower half of words was easier to read.
    }
    \label{fig:ease_rating}
\end{figure}

As shown in Figure \ref{fig:ease_rating}, in both Chinese and English, participants overwhelmingly rated the upper half of words as easier to read. This asymmetry was more pronounced in English, where 91\% of participants preferred the upper half, compared to 75\% in Chinese.

\section{Human Performance on Comprehension Questions}
\label{app:RCQ_performace_table}

\begin{table}[ht]
\centering
\begin{tabular}{llccc}
\toprule
\textbf{Language} &  & \textbf{Full} & \textbf{Upper} & \textbf{Lower} \\
\midrule
Chinese & Acc      & 66\% & 60\% & 56\% \\
English & Acc      & 81\% & 77\% & 60\% \\
\bottomrule
\end{tabular}
\caption{Comprehension question accuracy (Acc) 
for Chinese and English participants. 
}
\label{tab:rcq_acc}
\end{table}

We did not include these results from Table \ref{tab:rcq_acc} in the main text due to limitations in our experimental design. In the Chinese experiment, some screens lacked questions due to paragraph splitting, leading to mismatches between question accuracy and occlusion condition, i.e., answers for a given question could appear on the previous screen under a different occlusion condition. This likely explains the lower overall accuracy in Chinese. In the English experiment, we corrected this issue by including questions for all screens, though they were not generated with the original OneStopQA procedure \citep{berzak-etal-2020-starc} due to resource constraints.

\end{document}